\documentclass[11pt,letterpaper]{article}
\usepackage{authblk}
\usepackage{emnlp2017}

\usepackage{times}
\usepackage{latexsym}
\usepackage{multirow}
\usepackage{amsmath}
\usepackage{textcomp}
\usepackage{graphicx}
\usepackage{url}
\usepackage{subfigure}
\usepackage{enumitem}
\usepackage{arabtex}
\usepackage{utf8}
\setcode{utf8}
\emnlpfinalcopy

\usepackage{array}
\newcolumntype{L}[1]{>{\raggedright\let\newline\\\arraybackslash\hspace{0pt}}m{#1}}
\newcolumntype{C}[1]{>{\centering\let\newline\\\arraybackslash\hspace{0pt}}m{#1}}
\newcolumntype{R}[1]{>{\raggedleft\let\newline\\\arraybackslash\hspace{0pt}}m{#1}}

\title{Arabic Multi-Dialect Segmentation: bi-LSTM-CRF vs. SVM}

\author[1]{\bf Mohamed Eldesouki}
\author[2]{\bf Younes Samih}
\author[1]{\\ \bf Ahmed Abdelali}
\author[3]{\bf Mohammed Attia}
\author[1]{\bf Hamdy Mubarak}
\author[1]{\bf Kareem Darwish}
\author[2]{\bf Laura Kallmeyer}

\affil[1]{Qatar Computing Research Institute, HBKU, Doha, Qatar}
\affil[2]{Dept. of Computational Linguistics,University of D\"usseldorf, D\"usseldorf, Germany}
\affil[3]{Google Inc., New York City, USA}

\affil[1]{\texttt{\{mohamohamed,hmubarak,aabdelali,kdarwish\}@hbku.edu.qa}}
\affil[2]{\texttt{\{samih,kallmeyer\}@phil.hhu.de} }
\affil[3]{\texttt{attia@google.com} }

\date{}
\begin{document}
\maketitle

\begin{abstractx}
% Abstract \<والعربية اللغة>
Arabic word segmentation is essential for a variety of NLP applications such as machine translation and information retrieval. Segmentation entails breaking words into their constituent stems, affixes and clitics. In this paper, we compare two approaches for segmenting four major Arabic dialects using only several thousand training examples for each dialect.  The two approaches involve posing the problem as a ranking problem, where an SVM ranker picks the best segmentation, and as a sequence labeling problem, where a bi-LSTM RNN coupled with CRF determines where best to segment words. We are able to achieve solid segmentation results for all dialects using rather limited training data. We also show that employing Modern Standard Arabic data for domain adaptation and assuming context independence improve overall results.
\end{abstractx}

\section{Introduction}
Arabic has both complex morphology and orthography, where stems are typically derived from a closed set of roots to which affixes such as coordinating conjunctions, determiners, and pronouns are attached to form words. Segmenting Arabic words into their constituent parts is important for a variety of natural language processing applications. For example, segmentation has been shown to improve the effectiveness of information retrieval \cite{darwish2014arabic} and machine translation \cite{habash2006arabic}. Most previous work has mostly focused on segmenting Modern Standard Arabic (MSA) achieving segmentation accuracies of nearly 99\% \cite{abdelali2016farasa,pasha2014madamira}. MSA is the lingua franca of the Arab world, and it is typically used in written and formal communications. Dialectal Arabic (DA) segmentation on the other hand has received limited attention, with most of the work focusing on the Egyptian dialect \cite{habash2013morphological,samih2017neural}. Arabic dialects are typically spoken and are used in informal communications.  The advent of the social media and the ubiquity of smart phones has led to a greater need for dialectal processing such as dialect identification \cite{eldesouki2016qcri,khurana2016multi}, morphological analysis \cite{habash2013morphological} and machine translation \cite{sennrich-haddow-birch:2016:P16-12,Sajjad:2013ac}.  Yet, dialectal training corpora for a variety of NLP modules, including segmentation, continue to be limited and often nonexistent.

In this work, we focus on the segmentation of four major Arabic dialects, namely Egyptian, Levantine, Gulf, and Maghrebi.  We particularly focus on DA text from Twitter, a popular social media platform, from which we can obtain large amounts of text in different dialects written by ordinary social media users and exhibiting nonstandard orthography. We employ two machine learning approaches for building robust segmentation modules using limited training data (350 tweets containing several thousand words per dialect).  In one approach, we pose the segmentation as a ranking problem where all possible segmentations of a word are ranked using a Support Vector Machine (SVM) based ranker. In the second, we use bidirectional Long Short Term Memory (bi-LSTM) Recurrent Neural Network (RNN) with Conditional Random Fields (CRF) to perform sequence labeling over the characters in words.  For both, we adopt the simplifying assumption that word segmentation can be reliably performed independent of context.  Though the assumption is not always correct, it has been shown to be fairly robust for more than 99\% of word occurrences in Arabic text \cite{abdelali2016farasa}.  Lastly, given the large overlap between MSA and DA, we employ segmented MSA data to further improve dialectal segmentation.

The contribution of this paper are as follows:
\begin{itemize}[leftmargin=*]
\setlength\itemsep{-0.3em}
\item We present robust DA segmenters for four major Arabic dialects.  We plan to open-source all of them.
\item We provide an exposition of challenges associated with performing in situ DA segmentation including segmentation guidelines and the effect of orthographic standardization. 
\item We compare two machine learning approaches that can generalize well even when limited training data is available.
\end{itemize}

% The Arabic language has various dialects and variants that exist in a continuous spectrum. They are a result of an interweave between the Arabic language that spread throughout the Middle East and North Africa and the indigenous languages in different countries. With the passage of time and the juxtaposition of cultures, dialects and variants of Arabic evolved and mutated. Among the varieties of Arabic, so-called Modern Standard Arabic (MSA) is the lingua franca of the Arab world, and it typically used in written and formal communications. On the other hand, Arabic dialects, such as Egyptian and Levantine, are usually spoken and used in informal communications.
%, especially on social networks such as Twitter and Facebook.
% As a highly inflected and agglutinated language, \emph{word segmentation} is one of the most important processing steps for Arabic and its Dialects. Word segmentation is considered an integral part for many higher Arabic NLP tasks such as part-of-speech tagging, parsing and machine translation. 

% For example, the Egyptian DA word \<ومكتبهاش> ``wmktbhA\$'' %\textcolor{red}{(Hamdy: I modified the example!)}  meaning: ``and he didn't write it'') includes five clitics ``w+m+ktb+hA+\$'', namely the coordinate conjunction ``w'', the negation prefix ``m'', the verb (stem) ``ktb'', the object pronoun ``hA'' as suffix, and the negation suffix ``\$''.

\section{Related Work}
Work on dialectal Arabic is fairly new compared to MSA. A number of research projects were devoted to dialect identification \cite{Biadsy:2009:SAD:1621774.1621784,Zbib:2012:MTA:2382029.2382037,zaidan2014arabic,eldesouki2016qcri}. There are five major dialects including Egyptian, Gulf, Iraqi, Levantine and Maghribi. Few resources for these dialects are available such as the CALLHOME Egyptian Arabic Transcripts (LDC97T19), which was made available for research as early as 1997. Newly developed resources include the corpus developed by \newcite{Bouamor14MultidialectalCorpus}, which contains 2,000 parallel sentences in multiple dialects and MSA as well as English translation.\\
For the segmentation,
\newcite{mohamed2012annotating} built a segmenter based on memory-based learning. The segmenter has been trained on a small corpus of Egyptian Arabic comprising 320 comments containing 20,022 words from \url{www.masrawy.com} that were segmented and annotated by two native speakers. They reported a 91.90\% accuracy on the task of segmentation. MADA-ARZ \cite{habash2013morphological} is an Egyptian Arabic extension of the Morphological Analysis and Disambiguation of Arabic (MADA). They trained and evaluated their system on both Penn Arabic Treebank (PATB) (parts 1-3) and the Egyptian Arabic Treebank (parts 1-5) \cite{maamouri2014developing} and they achieved 97.5\% accuracy. MADAMIRA\footnote{MADAMIRA release 20160516 2.1} \cite{pasha2014madamira} is a new version of MADA and includes functionality for analyzing dialectal Egyptian. %  MADA-ARZ will be used in this paper for comparison.
\newcite{monroe2014word} used a single dialect-independent model for segmenting all Arabic dialects including MSA. They argue that their segmenter is better than other segmenters that use sophisticated linguistic analysis. They evaluated their model on three corpora, namely parts 1-3 of Penn Arabic Treebank (PATB), Broadcast News Arabic Treebank (BN), and parts 1-8 of the BOLT Phase 1 Egyptian Arabic Treebank (ARZ) reporting a 95.13\% F1 score.

\section{Dialectal Arabic}
\subsection{DA Challenges}
DA shares many MSA challenges, such as having complex templatic derivational morphology and concatenative orthography. Most  nouns and verbs are typically derived from a closed set of roots, which are fitted into templates to generate stems. Templates may indicate morphological features such POS tag, gender, and number. Stems may accept prefixes, such as coordinating conjunction and prepositions, or suffixes, such as pronouns, to form words. While dialects mostly comply with the templatic nature of morphology\footnote{Minor exception exist such the Egyptian template ``AtfEl'' that occasionally replaces the MSA template ``AnfEl'' as in \<اتكسر> ``Atksr'' (broke)}, they diverge from MSA in other aspects such as:
% and dialects introduce additional prefixes and suffixes and are more likely to concatenate or fuse words together.
% which turn dialectal Arabic into agglutinative and highly inflected language. Prefixes include coordinating conjunctions, determiner, and prepositions, and suffixes include attached pronouns and gender and number markers. This results in a large number of words (or surface forms) and in turn a high-level of sparseness and a large number of unseen words during testing.
% In addition to the shared challenges, dialectal Arabic has its own peculiarities. These peculiarities could be summarized as follows:
\begin{itemize}[leftmargin=*]
\setlength\itemsep{-0.3em}
\item Lack of standard orthography, particularly for strictly dialectal words such as \<عشان> ``E\$An'' (because), which may also appear as \<علشان> ``El\$An'' or \<مشان> ``m\$An'' \cite{habash2012conventional}. %. Many of the words in dialectal Arabic don't follow a standard orthographic system \cite{habash2012conventional}.
\item Word borrowing from other languages \cite{ibrahim2006borrowing}, such as \<بلاصتك> ``blAStk'' (your place) in Maghrebi, or code switching with other languages \cite{samih45676WSCS}.
\item Fusing multiple words together by concatenating tokens and dropping letters, such as the word \<يقولك> ``yqwlk'' (he says to you), ``yqwl lk'' are concatenated and one ``l'' is dropped.
\item Additional affixes.  Dialectal-specific affixes may arise because of: the alteration of pronouns, such as the feminine second person pronoun from \<ك> ``k'' to \<كي>``ky'' or the plural pronoun \<تم> ``tm'' to \<تو> ``tw''; the introduction of negation prefix-suffix combination \<ما> - \<ش> ``mA-\$'', which behaves like the French ``ne-pas'' negation construct; the placement of present tense markers, such as ``b'' in Egyptian and Levantine; the use of different future markers such as ``H'', ``h'', and ``g'' instead of ``s'' for MSA; and the shortening of prepositions and fusing them with the words they precede such as the transformation of \<على> ``ElY'' (on) to \<ع> ``E''.
\item Letter substitution, where some letters are commonly substituted for others such as ``v'' which is replaced with ``t'' in Egyptian (as in \<كتير> ``ktyr'' (much)) or ``q'' which is replaced with ``j'' in Gulf (as in \<صدج> ``Sdj'' (really).  %``D'' which is substituted with ``Z'' (as in \<ظابط> ``ZAbT'' (officer)). %, and ``Q'' and ``k'' which are substituted with ``g'' (as in \<صدج> (really) and \<يجذب> (lying), respectively).
\item Syntactic differences, such as the use of masculine plural or singular noun forms instead dual and feminine plural, the dropping of some articles and preposition in some syntactic constructs, and the abandonment of some suffixes such as ``wn'' in favor of ``wA'' for verbs and ``yn'' for nouns.
\end{itemize}

Using raw text from social media introduces additional phenomena such as word elongation, such \<أخييييررا> ``$>$KyyyyrrA'' (finally) instead of \<أخيرا> ``$>$KyrA'', and the use of non-Arabic characters such as Urdu characters \cite{darwish2012language}.
% In this paper, we present dialectal Egyptian segmentation using two approaches, namely Support Vector Machine (SVM$^{Rank}$) and Bidirectional Long-Short-Term-Memory (bi-LSTM).

% In this paper, we present dialectal Egyptian segmentation using Bidirectional Long-Short-Term-Memory (bi-LSTM).

\section{Dataset}
\label{sec:dataset}
  We constructed our dataset by obtaining 350 tweets that were authored for each of the following four dialects: Egyptian, Levantine, Gulf, and Maghrebi. For dialectal Egyptian tweets, we obtained the dataset described in \cite{darwish2014verifiably}, and we used the same methodology to construct the dataset for the remaining dialects. Initially, we obtained 175 million Arabic tweets by querying the Twitter API using the query ``lang:ar'' during March 2014.  Then, we identified tweets whose authors identified their location in countries where the dialects of interest are spoken (ex. Morocco, Algeria, and Tunisia for Maghrebi) using a large location gazetteer \cite{mubarak2014using}.  Then we filtered the tweets using a list containing 10 strong dialectal words per dialect, such as the Maghrebi word \<كيما> ``kymA'' (like/as in) and the Leventine word \<هيك> ``hyk'' (like this).  Given the filtered tweets, we randomly selected 2,000 unique tweets for each dialect, and we asked a native speaker of each dialect to manually select 350 tweets that are heavily dialectal.  Table \ref{table:annotatedData} lists the number of tweets that we obtained for each dialect and the number of words they contain.  
% which they tested in dialect identification task to distinguish between dialectal Egyptian and MSA. It contains 350 tweets with more than 8,000 words including 3,000 unique words written in Egyptian dialect. The tweets have much dialectal content covering most of dialectal Egyptian phonological, morphological, and syntactic phenomena. It also contains Twitter-specific aspects of the text, such as \#hashtags, $@$mentions, emoticons and URLs.

% We manually annotated each word in this corpus to provide: CODA-compliant writing \cite{habash2012conventional}, segmentation, stem, lemma, and POS, also the corresponding MSA word, MSA segmentation, and MSA POS. We share the dataset\footnote{Dataset is available at www.hidden-for-blind-review.com} to reproduce test results and help in other tasks such as CODA'fication of dialectal text, dialectal POS tagging and dialect to MSA conversion. Table~\ref{annotations} shows an annotation example of the word \<بيقولك> ``byqwlk'' (he is saying to you).

\begin{table}[h]
\begin{center}
\begin{tabular}{|l|l|}
\hline \bf Field & \bf Annotation \\ \hline
Orig. word & \<بيقولك> ``byqwlk''\\
Meaning & he is saying to you \\
In situ Segm. & \<ك>+\<يقول>+\<ب> ``b+yqwl+k''\\
CODA & \<بيقول لك> ``byqwl lk''\\
CODA Segm. & \<ك>+\<ل> \<يقول>+\<ب> ``b+yqwl l+k''\\
% POS & PROG\_PART+V PREP+PRON\\
% Stem & \<يقول ل> ``yqwl l''\\ 
% lemma & \<قال ل> ``qAl l''\\ 
% MSA & \<يقول لك> ``yqwl lk''\\
% MSA Segm. & \<ك>+\<يقول ل>``yqwl l+k''\\
% MSA POS & V PREP+PRON\\
\hline
\end{tabular}
\end{center}
\caption{Egyptian annotation example}
\vspace{-1.5em}
\label{annotations}
\end{table}

\begin{table}[h]
\begin{center}
\begin{tabular}{c|c|c}
Dialect & No of Tweets & No of Tokens \\ \hline
Egyptian & 350 & 6,721 \\
Levantine & 350 &  6,648 \\
Gulf & 350 & 6,844 \\
Maghrebi & 350 & 5,495 \\ \hline
\end{tabular}
\label{table:annotatedData}
\caption{Dataset size for the different dialects}
\end{center}
\vspace{-.75em}
\end{table}
% \textcolor{red}{Hamdy: We have exactly 350 tweets in all dialects. Right? Does "No of Tokens" contain also punct, numbers, URL's, etc.?}

% \subsection{Segmentation Conventions}
Segmentation of DA can be applied on the original raw text, or on the cleaned text after correcting spelling mistakes and applying conventional orthography rules, such as CODA \cite{habash2012conventional}.  In this work, we decided to segment the original raw text.   Though Egyptian CODA is a reasonably stable standard, CODA for other dialects are either immature or nonexistent. Also, CODA conversion tools are lacking for most dialects\footnote{except for Egyptian CODA tool that is embedded in MADAMIRA}. Building such tools requires the establishment of clear guidelines, is laborious, and may require large annotated corpora \cite{eskander2013processing}, such as the LDC Egyptian Treebank. 

To prepare the ground truth data for a dialect, we enlisted an annotator who is either a native speaker for the dialect or well versed in it and has background in natural language processing.  The authors along with another native speaker of the dialect made multiple review rounds on the work of the annotator to ensure consistency and quality.  
The annotation guidelines were fairly straightforward. Basically, we asked annotators to:
\begin{itemize}[leftmargin=*]
\setlength\itemsep{-0.3em}
\item segment words in a way that would maintain the correct number of part of speech tags
\item favor stems when repeated letters are dropped as Table \ref{annotations}
\item segment multiple concatenated words with pluses as in the ``merged words'' example in Table \ref{OriginalCODASegmentation}.
\item attach injected long vowels that trail prepositions or pronouns to the preposition or pronoun respectively (ex. \<ليكي> ``lyky'' (to you -- feminine) $\rightarrow$ ``ly+ky'')
\item treat dialectal words that originated as multiple fused words as single tokens (ex. \<علاش> ``ElA\$'' (why) -- originally \<على أي شيء> ``ElY $>$y \$y\textquotesingle '')
\item do not segment name mentions and hashtags
\end{itemize}

In what follows, we discuss the advantages and disadvantages of segmenting raw text versus the CODA'fied text with some statistics obtained for the Egyptian tweets for which we have a CODA'fied version as exemplified in Table~\ref{annotations}.  The main advantage of segmenting raw text is that it doesn't need any preprocessing tool to generate CODA orthography, and the main advantage of CODA is that is regularizes text making it more uniform and easier to process. 
% The best results they reported were 92.9\% after using Egyptian morphological analyzer. All these resources are not available for researchers, and they are either not existing or still not well established for other dialects.\newline
We manually compared the CODA version to the raw version of 2,000 words in our Egyptian dataset.  We found that in 75.4\% of the words, segmentation of original raw words is exactly the same as the their CODA'ifed equivalents (ex. \<و+من> ``w+mn'' (and from) and \<نعمل+ها>, ``nEml+hA'' (we do it)). Further, if we normalize some characters, namely \<أ،إ،آ$\leftarrow$ا، ى
$\leftarrow$ي، ة$\leftarrow$ه>, and non-Arabic characters \<چ$\leftarrow$ج، ڤ $\leftarrow$ف، گ،ک$\leftarrow$ك، ے$\leftarrow$ي>, and remove diacritics, the percentage of matching increases to 90.3\%. Table~\ref{OriginalCODASegmentation} showcases the remaining differences between raw and CODA segmentations and how often they appear.  The differences are divided into two groups.  In the first group (accounting for 6.8\% of the cases), the number of word segments remains the same and both the raw and CODA'fied segments would have the same POS tags.  

\begin{table}[h]
\begin{center}
\begin{tabular}{|l|l|l|}
\hline \bf Diff. & \textbf{\%} & \textbf{Examples} \\ \hline
\multicolumn{3}{|c|}{\textbf{Same no. of segments and same POS tags}} \\ \hline
variable	  & \multirow{2}{*}{2.4\%} & \<عشان $\Leftrightarrow$ علشان> \\
spellings &      & ``E\$An, El\$An''\\ \hline
% spellings &      & \<م+تعند+ش $\Leftrightarrow$ما تعند+ش>\\
%      &      & ``m+tEnd+\$, mA tEnd+\$''\\ \hline
dropped & \multirow{2}{*}{2.3\%} & \<ب+ حترم$\Leftrightarrow$ب+أحترم> \\
letters &      & ``b+Htrm, b+AHtrm''\\\hline
merged & \multirow{2}{*}{1.4\%}  &   \<يا+عم$\Leftrightarrow$يا عم> \\
words  &      & ``yA+Em, yA Em''\\\hline
Shortened  & \multirow{2}{*}{0.4\%}  &   \<ع$\Leftrightarrow$على، ف $\Leftrightarrow$في>\\
particles      &      &   ``E, ElA  f, fy''\\\hline
elongations & \multirow{1}{*}{0.3\%}  & \<لييييه$\Leftrightarrow$ليه> ``lyyyyh,lyh''             \\\hline
\multicolumn{3}{|c|}{\textbf{Different no. of segments or POS tags}} \\ \hline
spelling & \multirow{2}{*}{2.2\%}  &  \<انا$\Leftrightarrow$ان  ولا$\Leftrightarrow$وإلا>\\
errors &      & ``ِAn, Ana   wlA, wAlA''\\
\hline
fused & \multirow{2}{*}{0.8\%} & \<قال+ي$\Leftrightarrow$قال ل+ي>\\
letters &      & ``qAl+y, qAl l+y'' \\\hline
\end{tabular}
\end{center}
\caption{Original vs CODA Segmentations}
\label{OriginalCODASegmentation}
\end{table}

In this group, the ``variable spelling'' class contains dialectal words that may have different common spellings with one ``standard'' spelling in CODA. % are different than their CODA orthography which is a common case in DA as there is no standard writing, and many writings for the same word are accepted. This is similar to having ``thx'' and ``gr8'' as informal text.  writings for the standard English word "thanks". 
The ``dropped letter'' and ``shortened particles'' classes typically involve the omission of letters such as the first person imperfect prefix \<أ> ``$>$'' when preceded by the present tense marker \<ب> ``b'' or the future tense marker \<هـ> ``h'', and the ``A'' in negation particle \<ما>``mA'' which is often written as \<م>``m'' and attached to the following word, and the trailing letters in prepositions. % which is another common case in writing dialectal variations for words in negation constructs.
% The ``shortened particles'' contains mostly prepositions such as \<في، على> ``fy, ElY'' which are shortened to \<ف، ع> ``f, E''.  
``Merged words'' and ``word elongations'' are common in social media, where users try to keep within limit by dropping the spaces between letters that do not connect or to stress words respectively.%  particles \<يا، ما، لا> ``yA, mA, lA'' are merged to next word to form one word. In "ELONG" class, some letters are repeated to express some speech effects like "soooo much" in English.

Though some processing such as splitting of words or removing elongations is required to overcome the phenomena in this group, in situ segmentation of raw words would yield identical segments with the same POS tags as their CODA counterparts.  Thus, the segmentation of raw words could be sufficient for 97\% of words. 
% By applying minor modifications to next NLP components (for example adding some dialectal prefixes to the POS tagger), segmenting original words in classes "WRTG", "PREP", "MERGE", and "ELONG" can be considered as complex as segmenting CODAfied words. This increases the validation of the convention of segmenting original words to 95\%.

In the second group (accounting for 3\% of the cases), both may have a different number of segments or POS tags, which would complicate downstream processing such as POS tagging.  They involve spelling errors and the fusion of two identical consecutive letters (gemmination). Correcting such errors may require a spell checker.  We opted to segment raw input without correction in our reference, and we kept stem, such as verbs and nouns, complete at the expense of other segments such as prepositions as in the example in Table \ref{annotations}.

% In "ANIT:A" class, the beginning letter of present verbs \<أ> ``A'' indicating first person singular pronoun, is omitted when it's preceded by progressive or future particles \<ب، هـ> ``b, h''.

% In "L+PRON" case, the indirect object pronoun (prep \<ل>``l'' + direct object pronoun) is merged to previous word and they are written as one word, and the prep is dropped if the previous word ends with letter \<ل>``l''.

% In "ANIT:A" and "L+PRON" classes, although segmenting original words has good overlap with segmenting CODAfied words, it requires extra processing from next components to restore the missing letters in each class. This can be done by consulting a language model and/or a lexicon for each dialect which are not always available.

% In "SPELL" case, a word is misspelled (addition/deletion/substitution) such that the new generated form is still a valid word (most probably with different syntactic behavior). Correcting this kind of errors, requires contextual analysis and grammar checking which is not an easy task in DA and can be considered as a problematic case while segmenting either original or CODAfied words.

% In summary, segmenting original words is very close to segmenting CODAfied words (at least 95\%) and this ratio can be increased to 98\% by considering special predefined cases. It can be applied to raw text directly without the need of tools or lexicons for each dialect. These reported numbers here are for EGY dialect, and we plan to study other dialects for different segmentation conventions.

\section{Segmentation Approaches}
\label{segmentationapproaches}
We present here two different systems for word segmentation. The first uses SVM-based ranking (SVM$^{Rank}$)\footnote{\url{https://www.cs.cornell.edu/people/tj/svm_light/svm_rank.html}} to rank different possible segmentations for a word using a variety of features. The second uses bi-LSTM-CRF, which performs sequence-to-sequence mapping to guess word segmentation.

% In this section, we will provide a brief description of the bi-directional LSTM model.\\
% Brief description of the SVM model.

\subsection{SVM$^{Rank}$ Approach}
This approach is inspired by the work done by \newcite{abdelali2016farasa}, in which they used SVM based ranking to ascertain the best segmentation for Modern Standard Arabic (MSA), which they show to be fast and accurate. %Rather, it is an extension to the capability of Farasa to cover different Arabic dialects. 
The approach involves generating all possible segmentations of a word and then ranking them. 

In training, we generate all possible segmentations of a word based on a closed set of prefixes and suffixes, and the correct segmentation is assigned rank 1 and all other incorrect segmentations are assigned rank 2. Our valid affixes include MSA prefixes and suffixes that we extracted from Farasa \cite{abdelali2016farasa} and additional dialectal prefixes and suffixes that we observed during training.  Since we are not mapping words into a standard spelling, such as CODA, prefixes and suffixes may have multiple different representations.  For example, given the dialectal Egyptian word for ``I do not play'', it could be spelled as ``m+b+lEb+\$'', ``mA+b+lEb+\$'', ``mA+b+AlEb+\$'', ``m+b+lEb+\$y'', ``mA+b+AlEb+\$y'', etc.  In this example, the first prefix could be ``m'' or ``mA'' and the suffix could be ``\$'' or ``\$y''.  
% For the MSA affixes, we adopted the prefix and suffix lists from Farasa \cite{abdelali2016farasa}, namely:
% \begin{itemize}[leftmargin=*]
% \item Prefixes:  coordinating conjunctions (f, w), prepositions (l, b, k), determiner (Al), and future particle (s).
% \item Suffixes:  noun suffixes (p, At, An, wn, A, w, y, yn), case (A), and pronouns (t, k, n, wA, kmA, km, kn, h, hA, hmA, hm, hn, nA, tmA, tm, and tn).
% % Attia: What about verb suffixes?
% \end{itemize}	

% The prefixes are as follows:
% \begin{itemize}
% \item MSA prefixes: coordinating conjunctions (f, w), prepositions (l, b, k), determiner (Al), and future particle (s).
% \item Egyptian dialectal prefixes: future particle (h, H), preposition (E), and negation (m, mA).
% \end{itemize}

% % \textcolor{red}{Kareem $\rightarrow$ Desouki: how can ``E'' be a prefix? can you give me an example?}\\
% % \textcolor{red}{Desouki $\rightarrow$ Kareem: I found  two words in the data as \<عالفاضي> and \<عالماشي>}\\ 
% % The final set of prefixes would be:\\
% % (f, w, l, ll, b,  k,  Al,  s, h, H, E, m, mA)\\
% The suffixes are as follows:
% \begin{itemize}
% \item MSA suffixes: noun suffixes (p, At, An, wn, A, w, y, yn), case (A), and pronouns (t, k, n, wA, kmA, km, kn, h, hA, hmA, hm, hn, nA, tmA, tm, and tn).
% \item Egyptian dialectal suffixes: negation (\$), and pronouns (Ah, yA, ky, ny, ty, AhA, Ahm).
% \end{itemize}

Here are two example dialectal Egyptian words to demonstrate segmentation:
\begin{itemize}[leftmargin=*]
\setlength\itemsep{-0.3em}
\item Given the input word: \<عالوش> ``EAlw\$'' (on the face), possible segmentations are: \{E+Al+w\$\} (correct segmentation), \{E+Alw\$\}, \{E+Al+w+\$\}, \{E+Alw+\$\}, \{EAlw+\$\}, and \{EAlw\$\}.
\item Given the input word: \<باديكي> ``bAdyky'' (I give you (feminine)), possible segmentations are: \{b+Ady+ky\} (correct segmentation), \{b+Adyky\}, \{bAdy+ky\}, and \{bAdyky\}.
\end{itemize}

% When training SVM$^{Rank}$, we produce the feature vector for each possible segmentation, and we assign rank 1 to the correct segmentation and rank 2 for all other segmentations.

We use the following features in training the classifier:
\begin{itemize}[leftmargin=*]
\setlength\itemsep{-0.3em}
\item Conditional probability that a leading character sequence is a prefix.
\item Conditional probability that a trailing character sequence is a suffix.
\item probability of the prefix given the suffix.
\item probability of the suffix given the prefix.
\item unigram probability of the stem (more details about calculating this is showing below).
% \textcolor{red}{where is the probability coming from}
\item unigram probability of the stem with first suffix.
\item whether a valid stem template can be obtained from the stem.
\item whether the stem that has no trailing suffixes appears in a gazetteer of person and location names \cite{abdelali2016farasa}.
\item whether the stem is a function word, such as \<على> ``ElY'' (on), \<من> ``mn'' (from), and \<مش> ``m\$'' (not).
\item whether the stem appears in the AraComLex\footnote{\url{http://sourceforge.net/projects/aracomlex/}} Arabic lexicon \cite{attia2011open} or in the Buckwalter lexicon \cite{buckwalter2002buckwalter}. This is sensible considering the large overlap between MSA and DA.
\item length difference from the average stem length.
\end{itemize}

The segmentations with their corresponding features are then passed to the SVM ranker \cite{joachims2006training} for training. Our SVM$^{Rank}$ uses a linear kernel and a trade-off parameter between training error and margin of 100.

Before training the classifier, features needed to calculated in advance. As training data, we used the aforementioned sets of 350 dialectal tweets for each dialect containing typically several thousand words each. We also use three parts of the Penn Arabic Treebank (ATB); part 1 (version 4.1), 2 (version 3.1), and 3 (version 2), which have a combined size of 628,870 tokens, to lookup MSA segmentations.  The intuition behind using such segmented MSA data for lookup is that both MSA and dialects share a fair amount of vocabulary. Thus, using the ATB corpus has the effect of increasing coverage. % During training, we used either: the dialectal data only; or we combined both dialectal and MSA data.  When combining dialectal and MSA data, we duplicated the dialectal corpus multiple times until they reached about 400k words to nearly match the size of the ATB. The repeating assures the balance of the dialectal words when combined with the ATB corpus. The combined dialectal and MSA data is about 1 million words. 
% We used it to calculate the aforementioned feature values such as count of words, probability of prefixes, suffixes, prefixes given suffixes, suffixes given prefixes, and the average length of stem before generating the final features for the SVM$^{Rank}$.

We also adopted the simplifying assumption that any given word has only 1 possible correct segmentation regardless of context.  Though this assumption is not always true, previous work on MSA has shown that it holds for 99\% of the cases \cite{abdelali2016farasa}.  Invoking this assumption has multiple positive implications, namely: we can use the segmentations that we observed during training directly, which typically cover most common function words, or segmentations that we observed in the ATB, which cover most MSA words that may be prevalent in dialectal text; and we can cache word segmentations leading to significant speedup.  Thus, we experimented with three different lookup schemes for every word, namely: 1) we output the rankers guess directly (None); 2) if exists, we use seen segmentations in dialectal training set, and the output of the ranker otherwise (DA); 3) if exists, we use seen segmentation in dialectal training set, else we use segmentation that we observed in the ATB, and lastly the output of the ranker (DA+MSA).

% Another source of data was a set of collected dialectal tweets from Twitter. The tweets were authored during the period from March 2014 to April 2014. We obtained the tweets by posing the query "lang:ar" against Twitter API (using Twitter4j\footnote{\url{http://twitter4j.org/en/index.html}}) and filtering the tweets based on user declared location using a large gazetteer of user locations \cite{mubarak2014using}.  
% In all, we collected about 700k tweets of Egyptian Arabic. We used these tweets to compute the values of both {\em unigram probability of the stem} and {\em unigram probability of the stem with first suffix}.

\subsection{Bi-LSTM-CRF Approach}
\subsubsection{Long Short-term Memory}
Recurrent Neural Network (RNN) belongs to a family of neural networks suited for modeling sequential data. Given an input sequence $x = (x_1, ..., x_n)$, an RNN computes the output vector $y_t$ of each word $x_t$ by iterating the following equations from $t = 1$ to $n$:
\begin{align*}
h_t &= f(W_{xh} x_t + W_{hh} h_{t_{-1}} + b_h )\\
y_t &= W_{hy} h_t + b_y
\end{align*}
where $h_t$ is the hidden states vector, $W$ denotes weight matrix, $b$ denotes bias vector and $f$ is the activation function of the hidden layer. Theoretically, RNNs can learn long distance dependencies, still in practice they fail due vanishing/exploding gradients \cite{bengio:1994}.
To solve this problem , \newcite{hochreiter:1997} introduced the LSTM RNN. The idea consists of augmenting an RNN with memory cells to overcome difficulties with training and efficiently cope with long distance dependencies.
The output of the LSTM hidden layer $h_t$ given input $x_t$ is computed via the following intermediate calculations: \cite{graves:2013}:
\vspace*{-1mm}
\begin{align*}
\hspace*{5mm}
i_t &= \sigma(W_{xi} x_t + W_{hi} h_{t-1} + W_{ci} c_{t-1} + b_i )
\\ f_t &= \sigma(W_{xf} x_t + W_{hf} h_{t-1} + W_{cf} c_{t-1} + b_f )
\\c_t &= f_t c_{t-1} + i_t \tanh(W_{xc} x_t + W_{hc} h_{t-1} + b_c)
\\ o_t &= \sigma(W_{xo} x_t + W_{ho} h_{t-1} + W_{co} c_t + b_o )
\\ h_t &= o_t \tanh(c_t)
\end{align*}

where $\sigma$ is the logistic sigmoid function, and $i$, $f$, $o$ and $c$ are respectively the input gate, forget gate, output gate and cell activation vectors. More interpretation about this architecture can be found in \cite{Lipton:15}. 
%Figure~\ref{fig:lstm} illustrates a single LSTM memory cell \cite{graves:2005}

%\begin{figure}[h]
%\begin{center}
%\includegraphics[width=8cm,scale=0.4]{LSTM.png}
%\end{center}
%\caption{A Long Short-Term Memory Cell.}\label{fig:lstm}
%\end{figure}
\subsubsection{ Bi-directional LSTM}
\begin{figure}[t]
\begin{center}
\includegraphics[width=7.0cm,scale=0.5]{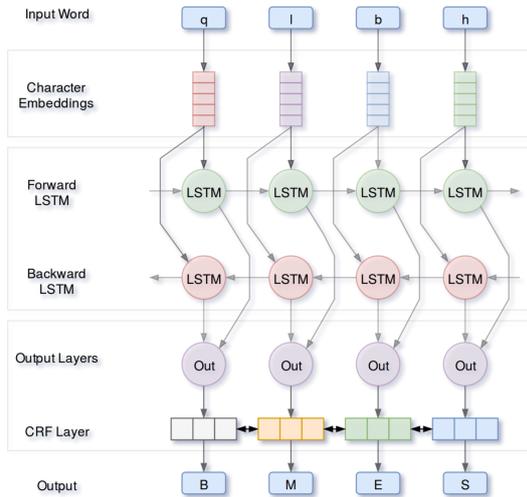}
\end{center}
\caption{Architecture of our proposed neural network Arabic segmentation model applied to word. \<قلبه> ``qlbh''  and output ``qlb+h'' }
\label{fig:seg_archit}
\vspace{-1.5em}
\end{figure}

Bi-LSTM networks \cite{schuster1997bilstm} are extensions to single LSTM networks. They are capable of learning long-term dependencies and maintain contextual features from past and future. As shown in Figure~\ref{fig:seg_archit}, they comprise two separate hidden layers that feed forward to the same output layer. A bi-LSTM calculates the forward hidden sequence $\overrightarrow{h}$, the backward hidden sequence $\overleftarrow{h}$ and the output sequence $y$ by iterating over the following equations :
\vspace*{-1mm}
\begin{align*}
\hspace*{5mm}
\overrightarrow{h_t} &= \sigma(W_{x\overrightarrow{h}} x_t + W_{\overrightarrow{h}\overrightarrow{h}} \overrightarrow{h}_{t-1}+ b_{\overrightarrow{h}})
\\ \overleftarrow{h_t} &= \sigma(W_{x\overleftarrow{h}} x_t + W_{\overleftarrow{h}\overleftarrow{h}} \overleftarrow{h}_{t-1}+ b_{\overleftarrow{h}})
\\ y_{t} &= W_{\overrightarrow{h y}}\overrightarrow{h_t} + W_{\overleftarrow{h y}}\overleftarrow{h_t} +
 b_y
 \end{align*}
More interpretations about these formulas are found in \newcite{graves2013hybrid}
\subsubsection{Conditional Random Fields (CRF)}
Over the past few years, bi-LSTMs have achieved many ground-breaking results in many NLP tasks because of their ability to cope with long distance dependencies and exploit contextual features from past and future states. Still, when they are used for some specific sequence classification tasks, (such as segmentation and named entity detection), where there is strict dependence between output labels, they fail to generalize perfectly. During the training phase of the bi-LSTM networks, the resulting probability distributions for different time steps are independent from each other. To overcome the independence assumptions imposed by the bi-LSTM and to exploit these kind of labeling constraints in our Arabic segmentation system, we model label sequence logic jointly using  Conditional Random Fields (CRF) ~\cite{crf}.
%A very important element of the recent success of many NLP applications, is the use of character-level representations in deep neural networks. This has shown to be effective for numerous NLP tasks \cite{collobert:2011b,dos:2015} as it can capture word morphology and reduce out-of-vocabulary. This approach has also been especially useful for handling languages with rich morphology and large character sets \cite{kim:2016}.
%We use pre-trained characters embeddings to initialize our look-up table. Unless otherwise specified, characters with no pre-trained embeddings are randomly initialized  with uniformly sampled embeddings. To use these embeddings in our model, we simply replace the one hot encoding character representation with its corresponding 200-dimensional vector. Table \ref{char_embb} shows the statistics of data we used to train our character embeddings. 

%\begin{table}[h]
%\begin{center}
%\begin{tabular}{|l|l|}
%\hline \bf Genre & \bf Tokens \\ \hline
%  Facebook posts & 8,241,244  \\ 
%         Tweets  & 2,813,016 \\
%  News comments  & 95,241,480 \\
% MSA news texts  & 276,965,735 \\ \hline
% \textbf{total}  & \textbf{383,261,475} \\ \hline
%\end{tabular}
%\end{center}
%\caption{character embeddings training data statistics}
%\label{char_embb}
%\end{table}

%We also find  this  important for the Arabic segmentation model particularly for our data as the MSA and DA have different orthographic sequences that are learned during the training phase.

\subsubsection{bi-LSTM-CRF for DA Segmentation}
%The architecture of our Arabic segmentation model, illustrated in Figure~\ref{fig:seg_archit}, is similar to 
In this model we consider Arabic segmentation as a sequence labeling problem at the character level. Each character is labeled with one of five labels ${B, M, E, S, WB}$ that designate the segmentation decision boundaries: Beginning, Middle, End of a multi-character segment, Single character segment, and Word Boundary respectively.  Figure \ref{fig:seg_archit} illustrates our segmentation model and how the model takes the word \<قلبه> ``qlbh'' (his heart) as its current input and predicts its correct segmentation. % The first layer performs a look up of the character embeddings and stacks them to build a matrix to input to the bi-LSTM. On the last layer, an affine transformation function followed by a CRF computes the probability distribution over all labels. 
% The architecture of our segmentation model, shown in Figure~\ref{fig:seg_archit}, is straightforward. 
The model is comprised of the following three layers:
\begin{itemize}[leftmargin=*]
\setlength\itemsep{-0.3em}
\item Input layer: containing character embeddings.
\item Hidden layer: bi-LSTM maps character representations to hidden sequences.
\item Output layer: CRF computes the probability distribution over all labels.
\end{itemize}

At the input layer, a look-up table is initialized with randomly uniform sampled embeddings\footnote{We did not use pre-trained character embeddings, because we conducted side experiments with and without pre-trained embeddings and the results were mixed} mapping each character in the input to d-dimensional vector. At the hidden layer, the output from the character embeddings is used as the input to the bi-LSTM layer to obtain fixed-dimensional representations for each character. At the output layer, a CRF is applied over the hidden representation of the bi-LSTM to obtain the probability distribution over all the labels.  Training is performed using stochastic gradient (SGD) descent with momentum $0.9$ and batch size $50$, optimizing the cross entropy objective function.

\subsubsection{Optimization}
Due to the relatively small size the training and development sets, overfitting poses a considerable challenge for our Dialectal Arabic segmentation system. To make sure that our model learns significant representations, we resort to dropout \cite{hinton:2012} to mitigate overfitting. The basic idea behind dropout involves randomly omitting a certain percentage of the neurons in each hidden layer for each presentation of the samples during training. This encourages each neuron to depend less on the other neurons to learn the right segmentation decision boundaries.
We apply dropout masks to the character embedding layer before inputting to the bi-LSTM and to its output vector. In our experiments, we find that dropout with a fixed rate of $0.5$ decreases overfitting and improves the overall performance of our system.
We also employ early stopping \cite{caruana2000,graves2013speech} to mitigate overfitting by monitoring the model's performance on the development set.

\begin{table*}[ht]
\begin{center}
\begin{tabular}{r|r|c|c|c|c}
Training Set & Look-up & Egyptian & Levantine & Gulf & Maghrebi \\ \hline
\multirow{3}{*}{SVM$^{Rank}$} & None 		& 91.0 & 87.8 & 87.7 & 84.7 \\
					& DA 		& 94.5 & 92.9 & 92.8 & 90.5 \\
					& DA+MSA 	& 94.6 & \textbf{93.3} & \textbf{93.1} & \textbf{91.2} \\ \hline
\multirow{3}{*}{bi-LSTM-CRF} & None 		& 93.8 & 91.0 & 89.4 & 87.1  \\
					& DA 		& 94.2 & 91.8 &90.8  & 88.5  \\
					& DA+MSA 	& \textbf{95.0} & 93.0 & 91.9 & 90.1  \\ \hline
\multicolumn{2}{c|}{Farasa}      & 85.7	& 82.6	& 82.9	& 82.6 \\ \hline

% \multirow{3}{*}{DA+MSA} & None 	& 90.5 & 88.3 & 87.4 & 84.9 \\
% 					& DA 		& 94.7 & 93.0 & 92.7 & 90.8 \\
% 					& DA+MSA 	& \bf{94.8} & \bf{93.4} & \bf{93.1} & \bf{91.4} \\ \hline
\end{tabular}
\label{SVMvbiLSTMCRFRes}
\caption{SVM$^{Rank}$ and bi-LSTM results %using dialectal or dialectal+MSA training data 
w/ and w/o lookup}
\end{center}
\vspace{-1.5em}
\end{table*}

\section{Experiments and Results}
% We used the entire dataset described in section~\ref{sec:dataset} for evaluating our SVM$^{rank}$ model. We used two evaluation metrics in our experiments, namely $F_1$ measure and accuracy. 

As described earlier, we perform several experiments for each dialect.  These involve training using dialectal data while using different lookup schemes, namely: no lookup (None); lookup from dialectal training only (DA); and a cascaded lookup from dialectal training and then MSA (DA+MSA).  For all our experiments, we use 5 fold cross validation with 70/10/20 train/dev/test splits.  % When using both DA and MSA training data, we placed all of the MSA data in the training split.
We use the Farasa MSA segmenter as a baseline. Table 4 % \ref{SVMvbiLSTMCRFRes} 
reports on the results for both segmentation approaches and in combination of using different lookup schemes.  As the results clearly shows, using an MSA segmenter yields suboptimal results for dialects.  Also, when no lookup is used, the bi-LSTM-CRF sequence labeler performs better than the SVM ranker for all dialects.  However, using lookup leads to greater improvements for the SVM approach leading to the best results for Levantine, Gulf, and Maghrebi and slightly lower results for Egyptian. Further, SVM$^{Rank}$ seemed to have benefited more from the DA lookup, while bi-LSTM-CRF benefited more from the MSA lookup. 
As for Egyptian segmentation, we suspected that it performed better for both approaches than the segmentation for the other dialects, because the percentage of test words that appear in the training set was greater for Egyptian.  The percentages for all the dialects are:
\begin{center}
\begin{tabular}{c|c|c|c}
Egyptian & Levantine & Gulf & Maghrebi \\ \hline
64.7\% & 54.7\% & 56.7\% & 55.2\% \\
% \hline
\end{tabular}
\end{center}
As for the lower results for Maghrebi, we noticed that Maghrebi has many more affixes than MSA and other dialects. These affixes contribute to the data sparsity and complexity of the segmentation task. For example, we enumerated 24 prefixes for Maghrebi compared to 8 for MSA, 17 for Levantine and Gulf, and 12 for Egyptian. Similarly, Maghrebi had more suffixes than MSA and other dialects. To ascertaining the effect of CODA'fication, we ran an extra experiment were we trained our best SVM$^{Rank}$ system using the CODA'fied version of the Egyptian data, and  the segmentation accuracy increased from 94.6\% to 96.8\%. Thus, having stable CODA standards and reliable conversion tools may positively impact dialectal processing. 
Next, we elaborate on typical errors of both approaches.\\

\noindent{\bf SVM$^{Rank}$ Errors:}  We examined the errors that the SVM ranker produced for different dialects and the most common types involved:
\begin{itemize}[leftmargin=*]
\setlength\itemsep{-0.3em}
\item erroneous splitting of leading or trailing characters when they were not prefixes or suffixes respectively or not splitting actual prefix and suffixes. For example, \<ح+يكون> ``H+ykwn'' (will be) was segmented as ``Hyk+wn''. % We suspect that such errors would go away with more training data as they typically involve stems that are unseen in training. 
\item the use of non-Arabic letters, wrong form of \textit{alef}, or ``h'' instead of ``p''.  For example \<جآي+ل+گ> ``jAy+l+k'' (I am coming to you), where ``A'' and ``k'' were replace with ``$|$'' and a Farsi character respectively, was not segmented.
\item long words with multiple segments such as \<م+تقلق+ي+نا+ش> ``m+tqlq+y+nA+\$'' (don't make us angry) where the ranker chose to segment it as ``m+tqlq+yn+A\$''.
\end{itemize}

\noindent{\bf bi-LSTM-CRF Errors:} The errors in this system are broadly classified into three categories:
\begin{itemize}[leftmargin=*]
\setlength\itemsep{-0.3em}
\item Ambiguous in token boundary because of character sharing in case of gemmination/elision. For example the word \<عنا> ``En$~$A'' (about us) is actually two tokens \<عن> En and \<نا> nA. The two \<ن> n letters are now merged into one. In the gold data, the disputed letter belongs to the first token while in the system output, it belongs to the second.
\item Like the SVM, the system often fails due to unconventional spelling. For example the word \<لاخويا> ``lAxwyA'' (to my brother) is a misspelling of \<لأخويا>. %, and therefore, the system fails to recognize the first letter as a token.
\item The majority of the remaining errors are simply mis-tokenization due to the system's inability to decide whether a substring (which out of context can be a valid token) is an independent token or part of a word, e.g. \<مستقبل+ك> ``mstqbl+k'' (your future), which is predicted by the system as ``m+staqbl+k'', where it correctly recognizes the genitive pronoun in the end, but mistakenly tags the first radical as a separate segment. % (probably confused with the negative/relative particle).
\end{itemize}

% Thus, both approaches produce similar types of errors.

\section{Conclusion}
In this paper we presented two approaches involving SVM-based ranking and bi-LSTM-CRF sequence labeling for segmenting Egyptian, Levantine, Gulf, and Maghrebi dialects. Both approaches yield strong comparable results that range between 91\% and 95\% accuracy for different dialects. To perform the work, we created training corpora containing naturally occurring text from social media for the aforementioned dialects.  We plan to release the data and the resulting segmenters to the research community. For future work, we want to perform domain adaptation using large MSA data, such as ATB, to improve segmentation results.  Further, we plan to investigate building a joint model capable of segmenting all the dialects with minimal loss in accuracy.

\bibliography{emnlp2017}

\begin{thebibliography}{}
\expandafter\ifx\csname natexlab\endcsname\relax\def\natexlab#1{#1}\fi

\bibitem[{Abdelali et~al.(2016)Abdelali, Darwish, Durrani, and
  Mubarak}]{abdelali2016farasa}
Ahmed Abdelali, Kareem Darwish, Nadir Durrani, and Hamdy Mubarak. 2016.
\newblock Farasa: A fast and furious segmenter for arabic.
\newblock In {\em Proceedings of the 2016 Conference of the North American
  Chapter of the Association for Computational Linguistics: Demonstrations\/}.
  Association for Computational Linguistics, San Diego, California, pages
  11--16.

\bibitem[{Attia et~al.(2011)Attia, Pecina, Toral, Tounsi, and van
  Genabith}]{attia2011open}
Mohammed Attia, Pavel Pecina, Antonio Toral, Lamia Tounsi, and Josef van
  Genabith. 2011.
\newblock An open-source finite state morphological transducer for modern
  standard arabic.
\newblock In {\em Proceedings of the 9th International Workshop on Finite State
  Methods and Natural Language Processing\/}. Association for Computational
  Linguistics, pages 125--133.

\bibitem[{Bengio et~al.(1994)Bengio, Simard, and Frasconi}]{bengio:1994}
Yoshua Bengio, Patrice Simard, and Paolo Frasconi. 1994.
\newblock Learning long-term dependencies with gradient descent is difficult.
\newblock {\em IEEE transactions on neural networks\/} 5(2):157--166.

\bibitem[{Biadsy et~al.(2009)Biadsy, Hirschberg, and
  Habash}]{Biadsy:2009:SAD:1621774.1621784}
Fadi Biadsy, Julia Hirschberg, and Nizar Habash. 2009.
\newblock Spoken arabic dialect identification using phonotactic modeling.
\newblock In {\em Proceedings of the EACL 2009 Workshop on Computational
  Approaches to Semitic Languages\/}. Association for Computational
  Linguistics, Stroudsburg, PA, USA, Semitic '09, pages 53--61.

\bibitem[{Bouamor et~al.(2014)Bouamor, Habash, and
  Oflazer}]{Bouamor14MultidialectalCorpus}
Houda Bouamor, Nizar Habash, and Kemal Oflazer. 2014.
\newblock A multidialectal parallel corpus of arabic.
\newblock In Nicoletta Calzolari~(Conference Chair), Khalid Choukri, Thierry
  Declerck, Hrafn Loftsson, Bente Maegaard, Joseph Mariani, Asuncion Moreno,
  Jan Odijk, and Stelios Piperidis, editors, {\em Proceedings of the Ninth
  International Conference on Language Resources and Evaluation (LREC'14)\/}.
  European Language Resources Association (ELRA), Reykjavik, Iceland.

\bibitem[{Buckwalter(2002)}]{buckwalter2002buckwalter}
Tim Buckwalter. 2002.
\newblock Buckwalter $\{$Arabic$\}$ morphological analyzer version 1.0 .

\bibitem[{Caruana et~al.(2000)Caruana, Lawrence, and Giles}]{caruana2000}
Rich Caruana, Steve Lawrence, and Lee Giles. 2000.
\newblock Overfitting in neural nets: Backpropagation, conjugate gradient, and
  early stopping.
\newblock In {\em NIPS\/}. pages 402--408.

\bibitem[{Darwish et~al.(2012)Darwish, Magdy, and Mourad}]{darwish2012language}
Kareem Darwish, Walid Magdy, and Ahmed Mourad. 2012.
\newblock Language processing for arabic microblog retrieval.
\newblock In {\em Proceedings of the 21st ACM international conference on
  Information and knowledge management\/}. ACM, pages 2427--2430.

\bibitem[{Darwish et~al.(2014{\natexlab{a}})Darwish, Magdy
  et~al.}]{darwish2014arabic}
Kareem Darwish, Walid Magdy, et~al. 2014{\natexlab{a}}.
\newblock Arabic information retrieval.
\newblock {\em Foundations and Trends{\textregistered} in Information
  Retrieval\/} 7(4):239--342.

\bibitem[{Darwish et~al.(2014{\natexlab{b}})Darwish, Sajjad, and
  Mubarak}]{darwish2014verifiably}
Kareem Darwish, Hassan Sajjad, and Hamdy Mubarak. 2014{\natexlab{b}}.
\newblock Verifiably effective arabic dialect identification.
\newblock In {\em EMNLP\/}. pages 1465--1468.

\bibitem[{Eldesouki et~al.(2016)Eldesouki, Dalvi, Sajjad, and
  Darwish}]{eldesouki2016qcri}
Mohamed Eldesouki, Fahim Dalvi, Hassan Sajjad, and Kareem Darwish. 2016.
\newblock Qcri@ dsl 2016: Spoken arabic dialect identification using textual
  features.
\newblock {\em VarDial 3\/} page 221.

\bibitem[{Eskander et~al.(2013)Eskander, Habash, Rambow, and
  Tomeh}]{eskander2013processing}
Ramy Eskander, Nizar Habash, Owen Rambow, and Nadi Tomeh. 2013.
\newblock Processing spontaneous orthography.

\bibitem[{Graves(2013)}]{graves:2013}
Alex Graves. 2013.
\newblock Generating sequences with recurrent neural networks.
\newblock {\em arXiv preprint arXiv:1308.0850\/} .

\bibitem[{Graves et~al.(2013{\natexlab{a}})Graves, Jaitly, and
  Mohamed}]{graves2013hybrid}
Alex Graves, Navdeep Jaitly, and Abdel-rahman Mohamed. 2013{\natexlab{a}}.
\newblock Hybrid speech recognition with deep bidirectional lstm.
\newblock In {\em Automatic Speech Recognition and Understanding (ASRU), 2013
  IEEE Workshop on\/}. IEEE, pages 273--278.

\bibitem[{Graves et~al.(2013{\natexlab{b}})Graves, Mohamed, and
  Hinton}]{graves2013speech}
Alex Graves, Abdel-rahman Mohamed, and Geoffrey Hinton. 2013{\natexlab{b}}.
\newblock Speech recognition with deep recurrent neural networks.
\newblock In {\em Acoustics, speech and signal processing (icassp), 2013 ieee
  international conference on\/}. IEEE, pages 6645--6649.

\bibitem[{Habash et~al.(2012)Habash, Diab, and Rambow}]{habash2012conventional}
Nizar Habash, Mona~T Diab, and Owen Rambow. 2012.
\newblock Conventional orthography for dialectal arabic.
\newblock In {\em LREC\/}. pages 711--718.

\bibitem[{Habash et~al.(2013)Habash, Roth, Rambow, Eskander, and
  Tomeh}]{habash2013morphological}
Nizar Habash, Ryan Roth, Owen Rambow, Ramy Eskander, and Nadi Tomeh. 2013.
\newblock Morphological analysis and disambiguation for dialectal arabic.
\newblock In {\em Hlt-Naacl\/}. pages 426--432.

\bibitem[{Habash and Sadat(2006)}]{habash2006arabic}
Nizar Habash and Fatiha Sadat. 2006.
\newblock Arabic preprocessing schemes for statistical machine translation.
\newblock In {\em Proceedings of the Human Language Technology Conference of
  the NAACL, Companion Volume: Short Papers\/}. Association for Computational
  Linguistics, pages 49--52.

\bibitem[{Hinton et~al.(2012)Hinton, Srivastava, Krizhevsky, Sutskever, and
  Salakhutdinov}]{hinton:2012}
Geoffrey~E Hinton, Nitish Srivastava, Alex Krizhevsky, Ilya Sutskever, and
  Ruslan~R Salakhutdinov. 2012.
\newblock Improving neural networks by preventing co-adaptation of feature
  detectors.
\newblock {\em arXiv preprint arXiv:1207.0580\/} .

\bibitem[{Hochreiter and Schmidhuber(1997)}]{hochreiter:1997}
Sepp Hochreiter and J{\"u}rgen Schmidhuber. 1997.
\newblock Long short-term memory.
\newblock {\em Neural computation\/} 9(8):1735--1780.

\bibitem[{Ibrahim(2006)}]{ibrahim2006borrowing}
Zeinab Ibrahim. 2006.
\newblock Borrowing in modern standard arabic.
\newblock {\em Innovation and Continuity in Language and Communication of
  Different Language Cultures 9. Edited by Rudolf Muhr\/} pages 235--260.

\bibitem[{Joachims(2006)}]{joachims2006training}
Thorsten Joachims. 2006.
\newblock Training linear svms in linear time.
\newblock In {\em Proceedings of the 12th ACM SIGKDD international conference
  on Knowledge discovery and data mining\/}. ACM, pages 217--226.

\bibitem[{Khurana et~al.(2016)Khurana, Ali, and Renals}]{khurana2016multi}
Sameer Khurana, Ahmed Ali, and Steve Renals. 2016.
\newblock Multi-view dimensionality reduction for dialect identification of
  arabic broadcast speech.
\newblock {\em arXiv preprint arXiv:1609.05650\/} .

\bibitem[{Lafferty et~al.(2001)Lafferty, McCallum, and Pereira}]{crf}
John~D. Lafferty, Andrew McCallum, and Fernando C.~N. Pereira. 2001.
\newblock Conditional random fields: {P}robabilistic models for segmenting and
  labeling sequence data.
\newblock In {\em Proc.~ICML\/}.

\bibitem[{Lipton et~al.(2015)Lipton, Kale, Elkan, and Wetzell}]{Lipton:15}
Zachary~C Lipton, David~C Kale, Charles Elkan, and Randall Wetzell. 2015.
\newblock A critical review of recurrent neural networks for sequence learning.
\newblock {\em CoRR\/} abs/1506.00019.

\bibitem[{Maamouri et~al.(2014)Maamouri, Bies, Kulick, Ciul, Habash, and
  Eskander}]{maamouri2014developing}
Mohamed Maamouri, Ann Bies, Seth Kulick, Michael Ciul, Nizar Habash, and Ramy
  Eskander. 2014.
\newblock Developing an egyptian arabic treebank: Impact of dialectal
  morphology on annotation and tool development.
\newblock In {\em LREC\/}. pages 2348--2354.

\bibitem[{Mohamed et~al.(2012)Mohamed, Mohit, and
  Oflazer}]{mohamed2012annotating}
Emad Mohamed, Behrang Mohit, and Kemal Oflazer. 2012.
\newblock Annotating and learning morphological segmentation of egyptian
  colloquial arabic.
\newblock In {\em LREC\/}. pages 873--877.

\bibitem[{Monroe et~al.(2014)Monroe, Green, and Manning}]{monroe2014word}
Will Monroe, Spence Green, and Christopher~D Manning. 2014.
\newblock Word segmentation of informal arabic with domain adaptation.
\newblock In {\em ACL (2)\/}. pages 206--211.

\bibitem[{Mubarak and Darwish(2014)}]{mubarak2014using}
Hamdy Mubarak and Kareem Darwish. 2014.
\newblock Using twitter to collect a multi-dialectal corpus of arabic.
\newblock In {\em Proceedings of the EMNLP 2014 Workshop on Arabic Natural
  Language Processing (ANLP)\/}. pages 1--7.

\bibitem[{Pasha et~al.(2014)Pasha, Al-Badrashiny, Diab, El~Kholy, Eskander,
  Habash, Pooleery, Rambow, and Roth}]{pasha2014madamira}
Arfath Pasha, Mohamed Al-Badrashiny, Mona Diab, Ahmed El~Kholy, Ramy Eskander,
  Nizar Habash, Manoj Pooleery, Owen Rambow, and Ryan~M Roth. 2014.
\newblock Madamira: {A} fast, comprehensive tool for morphological analysis and
  disambiguation of {A}rabic.
\newblock {\em Proc.~LREC\/} .

\bibitem[{Sajjad et~al.(2013)Sajjad, Darwish, and Belinkov}]{Sajjad:2013ac}
Hassan Sajjad, Kareem Darwish, and Yonatan Belinkov. 2013.
\newblock Translating dialectal {A}rabic to {E}nglish.
\newblock In {\em Proceedings of the 51st Annual Meeting of the Association for
  Computational Linguistics (Volume 2: Short Papers)\/}. Sofia, Bulgaria,
  ACL~'13, pages 1--6.

\bibitem[{Samih et~al.(2017)Samih, Attia, Eldesouki, Mubarak, Abdelali,
  Kallmeyer, and Darwish}]{samih2017neural}
Younes Samih, Mohammed Attia, Mohamed Eldesouki, Hamdy Mubarak, Ahmed Abdelali,
  Laura Kallmeyer, and Kareem Darwish. 2017.
\newblock A neural architecture for dialectal arabic segmentation.
\newblock {\em WANLP 2017 (co-located with EACL 2017)\/} page~46.

\bibitem[{Samih et~al.(2016)Samih, Maharjan, Attia, Kallmeyer, and
  Solorio}]{samih45676WSCS}
Younes Samih, Suraj Maharjan, Mohammed Attia, Laura Kallmeyer, and Thamar
  Solorio. 2016.
\newblock
  \href{http://www.aclweb.org/anthology/W/W16/W16-58.pdf\#page=62}{Multilingual
  code-switching identification via lstm recurrent neural networks}.
\newblock In {\em Proceedings of the Second Workshop on Computational
  Approaches to Code Switching,\/}. Austin, TX, pages 50--59.
\newblock
  \href{http://www.aclweb.org/anthology/W/W16/W16-58.pdf\#page=62}{http://www.aclweb.org/anthology/W/W16/W16-58.pdf\#page=62}.

\bibitem[{Schuster and Paliwal(1997)}]{schuster1997bilstm}
Mike Schuster and Kuldip~K Paliwal. 1997.
\newblock Bidirectional recurrent neural networks.
\newblock {\em IEEE Transactions on Signal Processing\/} 45(11):2673--2681.

\bibitem[{Sennrich et~al.(2016)Sennrich, Haddow, and
  Birch}]{sennrich-haddow-birch:2016:P16-12}
Rico Sennrich, Barry Haddow, and Alexandra Birch. 2016.
\newblock \href{http://www.aclweb.org/anthology/P16-1162}{Neural machine
  translation of rare words with subword units}.
\newblock In {\em Proceedings of the 54th Annual Meeting of the Association for
  Computational Linguistics (Volume 1: Long Papers)\/}. Association for
  Computational Linguistics, Berlin, Germany, pages 1715--1725.
\newblock
  \href{http://www.aclweb.org/anthology/P16-1162}{http://www.aclweb.org/anthology/P16-1162}.

\bibitem[{Zaidan and Callison-Burch(2014)}]{zaidan2014arabic}
Omar~F Zaidan and Chris Callison-Burch. 2014.
\newblock Arabic dialect identification.
\newblock {\em Computational Linguistics\/} 40(1):171--202.

\bibitem[{Zbib et~al.(2012)Zbib, Malchiodi, Devlin, Stallard, Matsoukas,
  Schwartz, Makhoul, Zaidan, and
  Callison-Burch}]{Zbib:2012:MTA:2382029.2382037}
Rabih Zbib, Erika Malchiodi, Jacob Devlin, David Stallard, Spyros Matsoukas,
  Richard Schwartz, John Makhoul, Omar~F. Zaidan, and Chris Callison-Burch.
  2012.
\newblock Machine translation of arabic dialects.
\newblock In {\em Proceedings of the 2012 Conference of the North American
  Chapter of the Association for Computational Linguistics: Human Language
  Technologies\/}. Association for Computational Linguistics, Stroudsburg, PA,
  USA, NAACL HLT '12, pages 49--59.

\end{thebibliography}


\begin{thebibliography}{}
\expandafter\ifx\csname natexlab\endcsname\relax\def\natexlab#1{#1}\fi

\bibitem[{Aho and Ullman(1972)}]{Aho:72}
Alfred~V. Aho and Jeffrey~D. Ullman. 1972.
\newblock {\em The Theory of Parsing, Translation and Compiling\/}, volume~1.
\newblock Prentice-Hall, Englewood Cliffs, NJ.

\bibitem[{{American Psychological Association}(1983)}]{APA:83}
{American Psychological Association}. 1983.
\newblock {\em Publications Manual\/}.
\newblock American Psychological Association, Washington, DC.

\bibitem[{Chandra et~al.(1981)Chandra, Kozen, and Stockmeyer}]{Chandra:81}
Ashok~K. Chandra, Dexter~C. Kozen, and Larry~J. Stockmeyer. 1981.
\newblock \href{https://doi.org/10.1145/322234.322243}{Alternation}.
\newblock {\em Journal of the Association for Computing Machinery\/}
  28(1):114--133.
\newblock
  \href{https://doi.org/10.1145/322234.322243}{https://doi.org/10.1145/322234.322243}.

\bibitem[{Gusfield(1997)}]{Gusfield:97}
Dan Gusfield. 1997.
\newblock {\em Algorithms on Strings, Trees and Sequences\/}.
\newblock Cambridge University Press, Cambridge, UK.

\end{thebibliography}
\bibliographystyle{emnlp_natbib}

\end{document}